\documentclass[sn-mathphys]{sn-jnl}

\usepackage[algo2e, ruled,linesnumbered, inoutnumbered]{algorithm2e}
\usepackage{paralist}
\usepackage{subfloat}
\usepackage{subfigure}
\newcommand{\etal}{\textit{et al}. }

\jyear{2023}%

\theoremstyle{thmstyleone}%
\theoremstyle{thmstyletwo}%

\theoremstyle{thmstylethree}%

\begin{document}

\title[Pre-trained weather model with masked autoencoder for weather forecasting]{W-MAE: Pre-trained weather model with masked autoencoder for multi-variable weather forecasting}


\author[1]{\fnm{Xin} \sur{Man}}

\author[2]{\fnm{Chenghong} \sur{Zhang}}

\author[3]{\fnm{Jin} \sur{Feng}}

\author[1]{\fnm{Changyu} \sur{Li}}

\author*[1]{\fnm{Jie} \sur{Shao}}\email{shaojie@uestc.edu.cn}

\affil[1]{\orgname{University of Electronic Science and Technology
of China}, \orgaddress{\city{Chengdu}, \country{China}}}

\affil[2]{\orgname{Institute of Plateau Meteorology, China
Meteorological Administration}, \orgaddress{\city{Chengdu},
\country{China}}}

\affil[2]{\orgname{Institute of Urban Meteorology, China
Meteorological Administration}, \orgaddress{\city{Beijing},
\country{China}}}


\abstract{Weather forecasting is a long-standing computational
challenge with direct societal and economic impacts. This task
involves a large amount of continuous data collection and exhibits
rich spatiotemporal dependencies over long periods, making it highly
suitable for deep learning models. In this paper, we apply
pre-training techniques to weather forecasting and propose W-MAE, a
Weather model with Masked AutoEncoder pre-training for weather
forecasting. W-MAE is pre-trained in a self-supervised manner to
reconstruct spatial correlations within meteorological variables. On
the temporal scale, we fine-tune the pre-trained W-MAE to predict
the future states of meteorological variables, thereby modeling the
temporal dependencies present in weather data. We conduct our
experiments using the fifth-generation ECMWF Reanalysis (ERA5) data,
with samples selected every six hours. Experimental results show
that our W-MAE framework offers three key benefits: 1) when
predicting the future state of meteorological variables, the
utilization of our pre-trained W-MAE can effectively alleviate the
problem of cumulative errors in prediction, maintaining stable
performance in the short-to-medium term; 2) when predicting
diagnostic variables (\textit{e.g.}, total precipitation), our model
exhibits significant performance advantages over FourCastNet; 3) Our
task-agnostic pre-training schema can be easily integrated with
various task-specific models. When our pre-training framework is
applied to FourCastNet, it yields an average 20\% performance
improvement in Anomaly Correlation Coefficient (ACC).}




\maketitle

\section{Introduction}

Weather forecasting is crucial for assisting in emergency
management, reducing the impact of severe weather events, avoiding
economic losses, and even generating sustained fiscal revenue
\cite{DBLP:journals/nature/BauerTB15}. Numerical Weather Prediction
(NWP) has emerged from applying physical laws to weather prediction
\cite{DBLP:journals/mwr/Abbe01, DBLP:journals/mz/Bjerknes09}. NWP
models \cite{DBLP:journals/pt/SchultzBGKLLMS21,
DBLP:journals/natmi/IrrgangBSBKSS21} use observed meteorological
data as initial conditions and perform numerical calculations with
supercomputers, solving fluid dynamics and thermodynamics equations
to predict future atmospheric movement states. Although modern
meteorological forecasting systems have achieved satisfactory
results using the NWP models, these models are subject to various
random factors due to their reliance on human understanding of
atmospheric physics, and may not meet the diverse forecasting needs
of complex climatic regions \cite{DBLP:journals/jmsj/Robert82}.
Moreover, numerical weather forecasting is a computation-intensive
task that requires the integration and analysis of large volumes of
diverse meteorological data, necessitating powerful computational
capabilities. Therefore, it is necessary to explore other
possibilities that can perform weather forecasting more
computational-efficiently and labor-savingly. Additionally, limiting
factors exist in NWP models, such as biases in parameterized
convection schemes that can seriously affect forecasting performance
\cite{DBLP:journals/pt/SchultzBGKLLMS21}. When using a single-model
forecast, in non-linear systems, tiny perturbations to the initial
field can lead to huge differences in results
\cite{DBLP:books/fv/03/TothTCZ03}. Ensemble forecast, which employs
models to produce a series of forecast results at the same forecast
time, can overcome this limitation of NWP
\cite{DBLP:journals/jcp/LeutbecherP08}. However, generating an
ensemble forecast with NWP models takes a long time, while
data-driven models can infer and generate forecasts several orders
of magnitude faster than traditional NWP models, thus enabling the
generation of very large ensemble forecasts
\cite{DBLP:journals/corr/abs-2103-09360}. Therefore, exploring the
performance and efficiency of models on ensemble forecasts is also
highly desirable.

In recent years, Artificial Intelligence (AI)-based weather
forecasting models using deep learning methods have attracted
widespread attention. Integrating deep learning methods into weather
forecasting requires a comprehensive consideration of the
characteristics of meteorological data and the advantages of
advanced deep learning technologies. Delving into the field of
meteorology, meteorological data involved in weather forecasting
exhibits characteristics such as vast data and lack of labels
\cite{DBLP:conf/nips/RacahBMKPP17, DBLP:journals/ijca/BiswasDB18},
diverse types \cite{DBLP:journals/cr/SemenovBBR98}, and strong
spatiotemporal dependencies \cite{DBLP:journals/ijon/CastroSOP021,
DBLP:conf/aaai/HanLZXD21}, which pose several challenges for the
AI-based weather forecasting models:
\begin{itemize}
\item \textbf{Massive unlabeled data mining}: Meteorological data is
commonly collected for monitoring and forecasting purposes, rather
than for targeted research or analysis. As a result, large amounts
of data accumulated over a long period of time may lack specific
labels or annotations about events or phenomena, which presents
challenges for effectively training and testing deep learning
models.
\item \textbf{Data assimilation \cite{Assimilation}}: Meteorological data
is not solely based on perceptual data but is numerical data that
integrates various sources of physical information and exhibits
diverse types. Therefore, weather forecasting requires the
integration of various meteorological data sources, which can be
noisy and heterogeneous. Many AI-based models
\cite{DBLP:journals/tjeecs/EsenerYK15,
DBLP:journals/remotesensing/AlbuCMCBM22} struggle to effectively
assimilate and learn from such diverse data.
\item \textbf{Spatial and temporal dependencies}: Meteorological data is
highly correlated and dependent in spatial and temporal
distributions due to the interplay and feedback mechanisms among
meteorological variables. Therefore, weather forecasting needs to
take these strong spatiotemporal dependencies into account. However,
most AI-based models \cite{DBLP:journals/remotesensing/DewitteCMM21,
DBLP:conf/ccc/DiaoNZC19} may not sufficiently capture these
dependencies, limiting their forecasting accuracy.
\end{itemize}

Pre-training techniques \cite{DBLP:conf/naacl/DevlinCLT19,
DBLP:conf/emnlp/HuangLDGSJZ19, DBLP:conf/aaai/LiDFGJ20} are
data-hungry, which is in line with the characteristic of large
amounts of weather data. Naturally, we introduce a self-supervised
pre-training technique into weather forecasting tasks. The
self-supervised pre-training technique
\cite{DBLP:journals/tcsv/TianGFFGFH22, DBLP:journals/tcsv/LiGCZLZ22}
aims to learn transferable representations from unlabeled data. It
utilizes intrinsic features of the data as supervisory signals to
automatically generate labels, obviating the need for manual data
annotation. The learning paradigm involves pre-training models on
large-scale unlabeled data, followed by fine-tuning for specific
downstream tasks. Through task-agnostic self-supervised
pre-training, on the one hand, models enable the direct utilization
of unlabeled data, expanding the available data range and quantity
without the need for manual data annotation; on the other hand,
pre-trained models can better handle heterogeneous data integration
by learning transferable representations.

To model the spatial and temporal dependencies in weather data, in
this study we employ the Vision Transformer (ViT) architecture
\cite{DBLP:conf/iclr/DosovitskiyB0WZ21} as the backbone network and
apply the widely-used pre-training scheme, Masked AutoEncoder (MAE)
\cite{DBLP:conf/cvpr/HeCXLDG22}, to propose a pre-trained weather
model named W-MAE for multi-variable weather forecasting. This
necessitates two considerations during the model-building process:

\begin{figure}[t]
\centering
\includegraphics[width=0.82\textwidth]{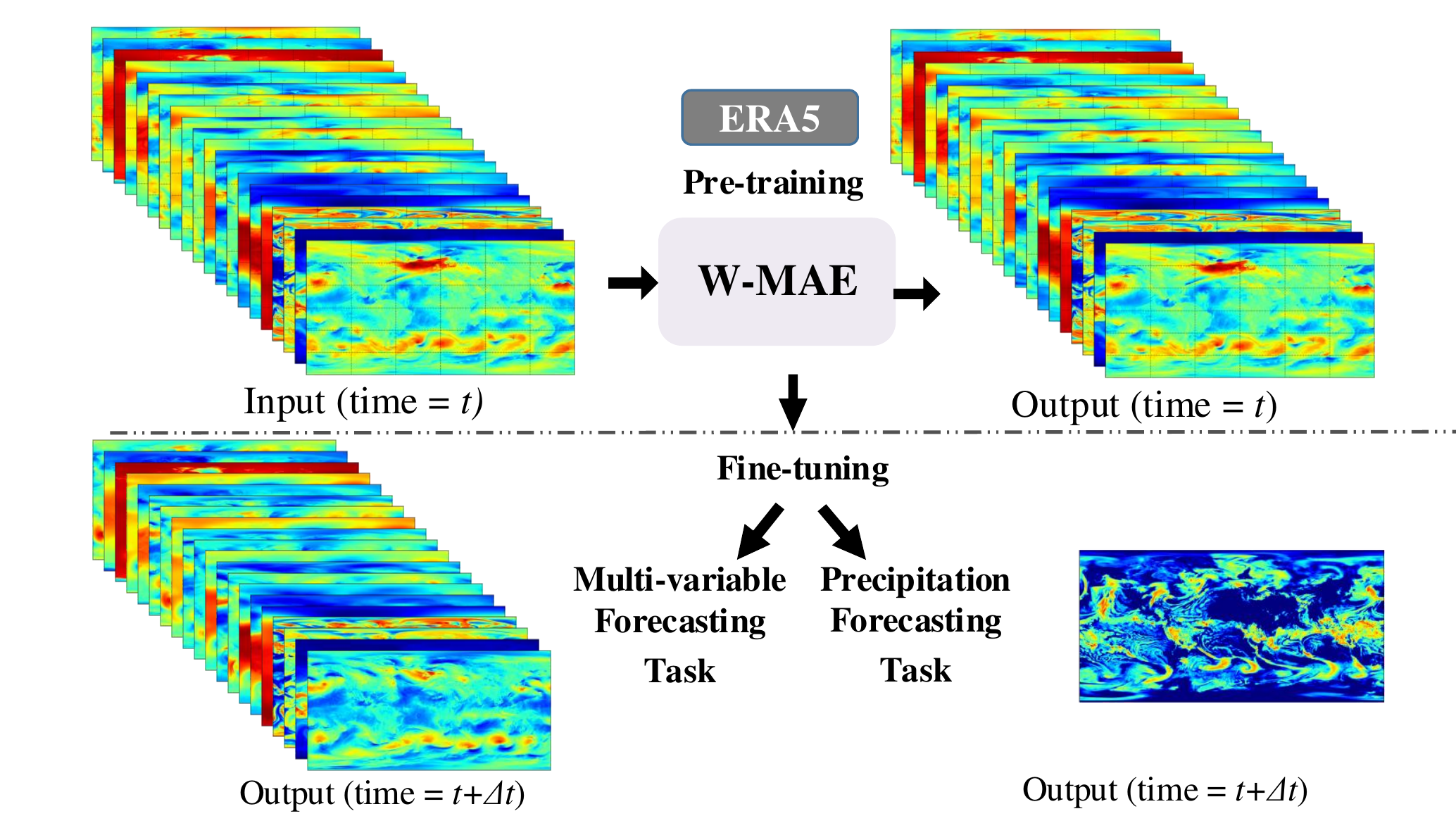}
\caption{A showcase of W-MAE pre-trained on the ERA5 dataset and
then fine-tuned on two weather forecasting tasks, \textit{i.e.},
multi-variable forecasting and precipitation forecasting. The word
in the grey boxes indicates the dataset used for pre-training and
fine-tuning our W-MAE.} \label{fig:showcase}
\end{figure}

\begin{itemize}
\item \textbf{Spatial dependency}: Unlike traditional computer vision data,
meteorological data is highly complex, and its interrelationships
are not straightforward. To address this challenge, we use MAE to
model twenty meteorological variables, each represented as a
two-dimensional pixel field of shape $721 \times 1440$, which
contains global latitude and longitude information
\cite{DBLP:journals/ames/RaspT21}. Notably, the size of the
two-dimensional pixel image of meteorological data is nearly three
times larger than the images ($224 \times 224$) processed in
traditional computer vision, which raises computational overheads.
To optimize the modeling ability of MAE while controlling
computational overheads, we modify the decoder structure by only
applying self-attention \cite{DBLP:conf/nips/VaswaniSPUJGKP17} to
the second dimension of the two-dimensional meteorological variables
image.
\item \textbf{Temporal dependency:} Considering the temporal dependencies
in meteorological data \cite{DBLP:journals/re/FaisalRHSHK22}, models
should learn from historical data and forecast future atmospheric
conditions. In this work, we fine-tune our pre-trained W-MAE to
predict future states of meteorological variables.
\end{itemize}

We choose the fifth-generation ECMWF Re-Analysis (ERA5) data
\cite{DBLP:journals/rms/HersbachBBH20} to establish our
meteorological data environment. Specifically, we select data
samples at six-hour intervals, where each sample comprises a
collection of twenty atmospheric variables across five vertical
levels. Figure~\ref{fig:showcase} shows an example of our W-MAE
pre-trained on the ERA5 dataset and then fine-tuned on multiple
downstream tasks.

Recent large-scale AI-based weather models such as FourCastNet
\cite{DBLP:journals/corr/abs-2202-11214}, GraphCast
\cite{DBLP:journals/corr/abs-2212-12794}, and Pangu-Weather
\cite{DBLP:journals/corr/abs-2211-02556} concentrate more on the
temporal scale and are typically task-specific. However, these
task-specific models often present a ``lazy'' behavior during
training, where they mainly prioritize the specific task
(\textit{i.e.}, forecasting for the next time-step), leading to
suboptimal performance on multi-step forecasting due to error
accumulation. To address this issue, FourCastNet
\cite{DBLP:journals/corr/abs-2202-11214} conducts a second round of
training to predict two time-steps, while GraphCast
\cite{DBLP:journals/corr/abs-2212-12794} increases the number of
autoregressive steps to 12 in its forecasting. Pangu-Weather
\cite{DBLP:journals/corr/abs-2211-02556} takes a different approach
by training four separate models, each specialized in forecasting
for different time intervals (\textit{i.e.}, 1, 3, 6, and 24 hours).
In practice, combining the forecasting results from these models can
yield optimal results with a very small number of iterations.

In contrast, our W-MAE involves reconstructing the masked pixels,
thereby modeling the spatial relationships within weather data. It
is worth noting that W-MAE is the first large-scale weather model
explicitly designed to consider the spatial dependencies within the
data. Besides, through task-agnostic pre-training, our W-MAE model
can learn from various proxy tasks that are independent of
downstream tasks. This allows our model to concentrate on exploring
the data itself, effectively leveraging massive amounts of unlabeled
meteorological data, and acquiring rich meteorological-related basic
features and general knowledge, thereby facilitating task-specific
fine-tuning. Experimental results demonstrate that our W-MAE model
achieves stable 10-day prediction performance for meteorological
variables (\textit{e.g.}, low-level winds and geopotential height at
500 hPa) without explicitly optimizing for error accumulation.
Furthermore, in the case of the more challenging diagnostic variable
forecasting, \textit{i.e.}, the precipitation forecasting task
\cite{DBLP:journals/rms/RodwellRHH10}, our W-MAE significantly
outperforms FourCastNet. These findings suggest that our
pre-training scheme enables models to provide valuable information
to downstream tasks, potentially mitigating error accumulation
problems to some extent and thereby achieving robust results and
performance improvements.

Moreover, our W-MAE model can be easily combined with task-specific
weather models to serve as a common basis for various weather
forecasting tasks. We apply our proposed pre-training framework to
FourCastNet and compare its performance under both pre-trained and
non-pre-trained settings. Experimental results show that with our
pre-training, FourCastNet exhibits a nearly 20\% improvement in the
task of precipitation forecasting. In terms of time overhead, direct
training of FourCastNet takes nearly 1.2 times longer than
fine-tuning a pre-trained FourCastNet. These findings highlight the
potential of our W-MAE architecture to enhance the model fine-tuning
phase, yielding significant time and resource savings while enabling
faster convergence and improved performance.

The main contributions of this work are three-fold:
\begin{itemize}
    \item We introduce self-supervised pre-training techniques into weather
forecasting tasks, enabling weather forecasting models fully mine
rich meteorological-related basic features and general knowledge
from large-scale, unlabeled meteorological data, and better handle
heterogeneous data integration.
    \item We employ the ViT architecture as the backbone network and apply the
widely-used pre-training scheme MAE to propose W-MAE, a pre-training
method designed for weather forecasting. Our pre-trained W-MAE
captures the underlying structures and patterns of the data,
providing powerful initial parameters for downstream task
fine-tuning. This facilitates faster convergence, resulting in more
robust and higher-performing models.
    \item Extensive experiments demonstrate that our model has a stable and
significant advantage in short-to-medium-range forecasting.
Additionally, we explore large-member ensemble forecasts and our
W-MAE model ensures fast inference speed and excellent performance
in a 100-member ensemble forecast, achieving high inference
efficiency.

\end{itemize}

\section{Related work}

In this section, we briefly review the development of weather
forecasting, as well as some technologies related to our work,
mainly involving numerical weather prediction, AI-based weather
forecasting and pre-training techniques.

\subsection{Numerical weather prediction}
\label{sec:nwp}

Numerical Weather Prediction (NWP) involves using mathematical
models, equations, and algorithms to simulate and forecast
atmospheric conditions and weather patterns
\cite{DBLP:journals/nature/BauerTB15}. NWP models generally use
physical laws and empirical relationships such as convection,
advection, and radiation to predict future weather states
\cite{DBLP:journals/msj/Arakawa97}. The accuracy of NWP models
depends on many factors, such as the quality of the initial
conditions, the accuracy of the physical parameterizations
\cite{Parameterization}, the resolution of the grid, and the
computational resources available. In recent years, many efforts
have been made to improve the accuracy and efficiency of NWP models
\cite{DBLP:journals/pt/SchultzBGKLLMS21,
DBLP:journals/natmi/IrrgangBSBKSS21}. For example, some researchers
\cite{DBLP:journals/jc/Arakawa04, DBLP:journals/mwr/GrenierB01} have
proposed using adaptive grids to better capture the features of the
atmosphere, such as the boundary layer and the convective clouds.
Others have proposed using hybrid models
\cite{DBLP:journals/ae/HanMSCLLX22} that combine the strengths of
different models, such as the global and regional models
\cite{DBLP:journals/geoinformatica/ChenWZZLY19}.

\subsection{AI-based weather forecasting}

In recent years, there has been an increasing interest in the
development of AI-based weather forecasting models. These models use
deep learning algorithms to analyze vast amounts of meteorological
data and learn patterns that can be used to make accurate weather
forecasting. Compared with the traditional NWP models
\cite{DBLP:journals/pt/SchultzBGKLLMS21,
DBLP:journals/natmi/IrrgangBSBKSS21}, AI-based models
\cite{DBLP:journals/ames/RaspDSWMT20,
DBLP:journals/corr/abs-2202-07575} have the potential to produce
more accurate and timely weather forecasts, especially for extreme
weather events such as hurricanes and heat waves. Albu \etal
\cite{DBLP:journals/remotesensing/AlbuCMCBM22} combine Convolutional
Neural Network (CNN) with meteorological data and propose NeXtNow, a
model with encoder-decoder convolutional architecture. NeXtNow is
designed to analyze spatiotemporal features in meteorological data
and learn patterns that can be used for accurate weather
forecasting. Karevan and Suykens \cite{DBLP:journals/nn/KarevanS20}
explore the use of Long Short-Term Memory (LSTM) network for weather
forecasting, which can capture temporal dependencies of
meteorological variables and are suitable for time series
forecasting, but may struggle to capture spatial features. However,
both LSTM-based and CNN-based models suffer from high computational
costs, limiting their ability to handle large amounts of
meteorological data in real-time applications.

Taking into account the computational costs and the need for timely
forecasting, Pathak \etal \cite{DBLP:journals/corr/abs-2202-11214}
propose FourCastNet, an AI-based weather forecasting model that
employs adaptive Fourier neural operators to achieve high-resolution
forecasting and fast computation speeds. FourCastNet represents a
promising solution for real-time weather forecasting, but it
requires significant amounts of training data to achieve optimal
performance and may have limited accuracy in certain extreme weather
events. In this work, we apply our pre-training method to
FourCastNet to explore its impact on model performance. The results
show that pre-trained FourCastNet achieves nearly 20\% performance
improvement in precipitation forecasting. This suggests that
pre-training can be a feasible strategy to enhance the performance
of FourCastNet and other weather forecasting models.

\subsection{Self-supervised pre-training techniques}

Self-supervised learning enables pre-training rich features without
human annotations, which has made significant strides in recent
years. In particular, Masked AutoEncoder (MAE)
\cite{DBLP:conf/cvpr/HeCXLDG22}, a recent state-of-the-art
self-supervised pre-training scheme, pre-trains a ViT encoder
\cite{DBLP:conf/iclr/DosovitskiyB0WZ21} by masking an image, feeding
the unmasked portion into a Transformer-based encoder, and then
tasking the decoder with reconstructing the masked pixels. MAE
adopts an asymmetric design that allows the encoder to operate only
on the partial, observed signal (\textit{i.e.}, without mask tokens)
and a lightweight decoder that reconstructs the full signal from the
latent representation and mask tokens. This design achieves a
significant reduction in computation by shifting the mask tokens to
the small decoder. Our W-MAE model is built upon the MAE
architecture and also utilizes the VIT encoder to process unmasked
image patches. However, we employ a modified decoder specifically
designed to reconstruct pixels for meteorological data to reduce
computational overhead.

\section{Preliminaries}

In this section, we provide a brief introduction to the ERA5 dataset
and downstream weather forecasting tasks, as foundations for our
subsequent presentation of the W-MAE model and training details.

\subsection{Dataset}
\label{sec:dataset}

ERA5 \cite{DBLP:journals/rms/HersbachBBH20} is a publicly available
atmospheric reanalysis dataset provided by the European Centre for
Medium-Range Weather Forecasts (ECMWF). The ERA5 reanalysis data
combines the latest forecasting models from the Integrated
Forecasting System (IFS) \cite{DBLP:journals/ams/BougeaultTBBB10}
with available observational data (\textit{e.g.}, pressure,
temperature, humidity) to provide the best estimates of the state of
the atmosphere, ocean-wave, and land-surface quantities at any point
in time. The current ERA5 dataset comprises data from 1979 to the
present, covering a global latitude-longitude grid of the Earth's
surface at a resolution of $0.25^{\circ} \times 0.25^{\circ}$ and
hourly intervals, with various climate variables available at 37
different altitude levels, as well as at the Earth's surface. Our
experiments are conducted on the ERA5 dataset across two distinct
data partition schemes. The first partition manifests a division
wherein the training, validation, and test sets are composed of data
from 2015, 2016 and 2017, and 2018, respectively. The second
partition arrangement aligns with the division used in FourCastNet
\cite{DBLP:journals/corr/abs-2202-11214} whereby the training,
validation, and test sets are comprised of data from 1979 to 2015,
2016 and 2017, and 2018, respectively. We select data samples at
six-hour intervals, where each sample comprises a collection of
twenty atmospheric variables across five vertical levels (see
Table~\ref{tab:variables} for details).

\begin{table}[t]
\centering
\caption{The abbreviations are as follows: $U_{10}$ and
$V_{10}$ represent the zonal and meridional wind velocity,
respectively, at a height of 10m from the surface; $T_{2m}$
represents the temperature at a height of 2m from the surface; $T$,
$V$, $Z$, and $RH$ represent the temperature, zonal velocity,
meridional velocity, geopotential, and relative humidity,
respectively, at specified vertical levels; and $TCWV$ represents
the total column water vapor.}
\resizebox{\textwidth}{!}{ }
\begin{tabular}{cc}
\toprule
Vertical level & Variables               \\
\midrule
Surface        & $U_{10}, V_{10}, T_{2m}, sp, mslp$ \\
10000 hPa       & $U, V, Z$                 \\
850 hPa         & $T, U, V, Z, RH$          \\
500 hPa         & $T, U, V, Z, RH$          \\
50 hPa          & $Z$                       \\
Integrated     & $TCWV$    \\
\bottomrule
\end{tabular}
\label{tab:variables}
\end{table}

\subsection{Multi-variable and precipitation tasks}

We focus on forecasting two important and challenging atmospheric
variables (consistent with the work done on FourCastNet
\cite{DBLP:journals/corr/abs-2202-11214}): 1) the wind velocities at
a distance of 10m from the surface of the earth and 2) the 6-hourly
total precipitation. The variables are selected for the following
reasons: 1) predicting near-surface wind speeds is of tremendous
practical value as they play a critical role in planning energy
storage and grid transmission for onshore and offshore wind farms,
among other operational considerations; 2) neural networks are
particularly well-suited to precipitation prediction tasks due to
their impressive ability to infer parameterizations from
high-resolution observational data. Additionally, our model reports
forecasting results for several other variables, including
geopotential height, temperature, and wind speed.

\section{Pre-training method}

\begin{figure}[t]
\centering
\includegraphics[width=0.9\textwidth]{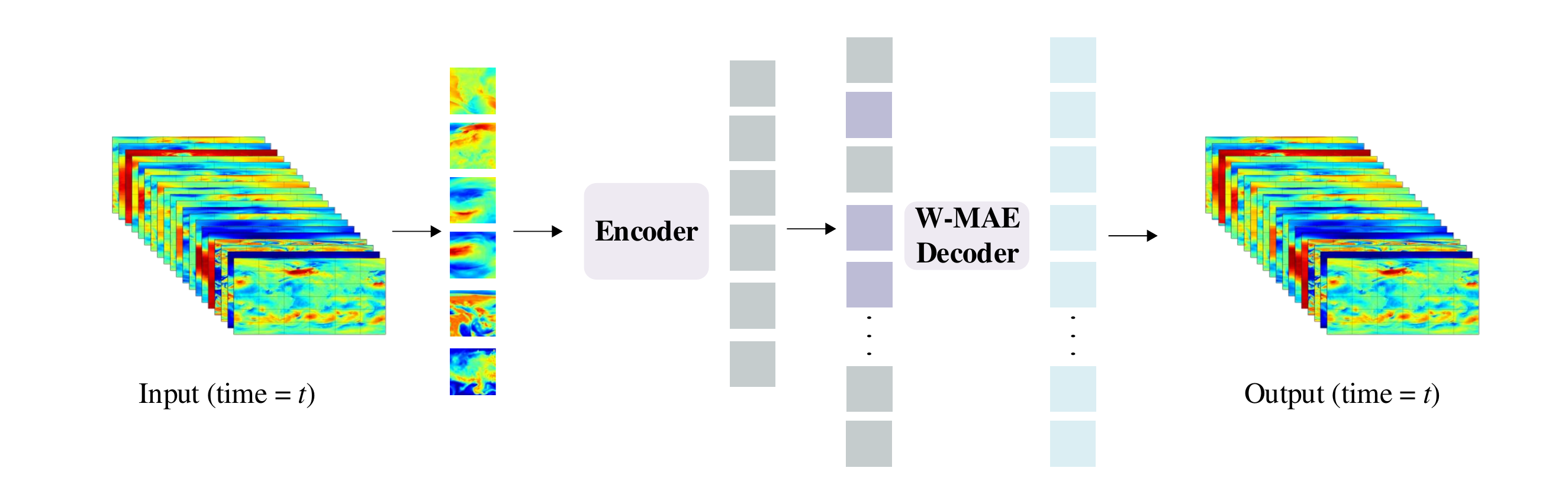}
\caption{Our proposed W-MAE architecture. Following the MAE
framework, an input image is first divided into patches and masked
before being passed into the encoder. The encoder only processes the
unmasked patches. After the encoder, the removed patches are then
placed back into their original locations in the sequence of patches
and fed into our W-MAE decoder to reconstruct the missing pixels.}
\label{fig:architecture}
\end{figure}

The pre-training task is to generate representative features by
randomly masking patches of the input meteorological image and then
reconstructing the missing pixels. In this section, we provide a
detailed description of our W-MAE architecture, as illustrated in
Figure~\ref{fig:architecture}. Our W-MAE employs the ViT
architecture as the backbone network and applies the MAE
pre-training scheme. Compared with the vanilla MAE, the proposed
W-MAE uses a modified decoder structure to save computation. We
formalize W-MAE in the following, by first specifying the necessary
VIT and MAE background and then explaining our W-MAE decoder.

\subsection{VIT and MAE}

Vision Transformer (ViT) \cite{DBLP:conf/iclr/DosovitskiyB0WZ21},
with its remarkable spatial modeling capabilities, scalability, and
computational efficiency, has emerged as one of the most popular
neural architectures. Let $I\in\mathbb{R}^{H\times W\times C}$
denote an input image with height $H$, width $W$, and $C$ channels.
In our study, we treat each sample of ERA5 data as an image, with
different channels representing the different atmospheric variables
contained in each sample. ViT differs from the standard Transformer
in that it divides an image into a sequence of two-dimensional
patches $\mathbf{x}_{p}\in\mathbb{R}^{N\times(P^{2}\cdot C)}$, where
each patch has a resolution of $(P, P)$, resulting in $N = HW/P^{2}$
patches. The patches are flattened and mapped to a $D$-dimensional
feature space using a trainable linear projection
(Eq.~\ref{eq:proj}). The output of this projection, referred to as
patch embeddings, is then concatenated with a learnable embedding,
which serves as the input sequence to the Transformer encoder. The
encoder consists of alternating layers of Multi-headed
Self-Attention (MSA) and Multi-Layer Perceptron (MLP) blocks
(Eq.~\ref{eq:msa} and Eq.~\ref{eq:mlp}). The MLP block contains two
fully connected layers with a GELU activation function. Layer
Normalization (LN) is applied before each block, and residual
connections are added. The overall procedure can be formalized as
follows:
\begin{equation}
\begin{aligned}
\label{eq:proj}
\mathbf{z}_0=[\mathbf{x}^1_p\mathbf{E};\mathbf{x}_{p}^2\mathbf{E};\cdots;\mathbf{x}_p^N\mathbf{E}]+\mathbf{E}_{pos},
\end{aligned}
\end{equation}
\begin{equation}
\begin{aligned}
\label{eq:msa}
\mathbf{z}'_\ell=\operatorname{MSA}(\operatorname{LN}(\mathbf{z}_{\ell-1}))+\mathbf{z}_{\ell-1},\quad&\ell=1\ldots
L,
\end{aligned}
\end{equation}
\begin{equation}
\begin{aligned}
\label{eq:mlp}
\mathbf{z}_\ell=\operatorname{MLP}(\operatorname{LN}(\mathbf{z'}_\ell))+\mathbf{z'}{}_\ell,\quad&\ell=1{\ldots
L},
\end{aligned}
\end{equation}
where $\mathbf{E}\in\mathbb{R}^{(P^2\cdot C)\times D}$,
$\mathbf{E}_{pos}\in\mathbb{R}^{(N+1)\times D}$ and $L$ is the
number of layers.

MAE is a self-supervised pre-training scheme that masks random
patches of the input image and reconstructs the missing pixels.
Specifically, a masking ratio of $m$ is selected, and $m$ percentage
of the patches is randomly removed from the input image. The
remaining patches are then passed through a projection function
(\textit{e.g.}, a linear layer) to project each patch (contained in the
remaining patch sequence $S$) into a $D$-dimensional embedding. A
positional encoding vector is then added to the embedded patches to
preserve their spatial information, and the function is defined as:
\begin{equation}
\begin{aligned}
\label{eq:vx} v_x(pos,2i)&=\sin\frac{pos}{10000^{\frac{2i}{D}}},
\end{aligned}
\end{equation}
\begin{equation}
\begin{aligned}
\label{eq:vy} v_y(pos,2i+1)&=\cos\frac{pos}{10000^{\frac{2i}{D}}},
\end{aligned}
\end{equation}
where $pos$ is the position of the patch along the given axis and
$i$ is the feature index. Subsequently, the resulting sequence is
fed into a Transformer-based encoder. The encoder processes the
sequence of patches and produces a set of latent representations.
The removed patches are then placed back into their original
locations in the sequence of patches, and another positional
encoding vector is added to each patch to preserve the spatial
information. After that, all patch embeddings are passed to the
decoder to reconstruct the original input image. The objective
function is to minimize the difference between the input image and
the reconstructed image. Our proposed W-MAE architecture utilizes
MAE to reconstruct the spatial correlations within meteorological
variables.

\subsection{W-MAE decoder}

Our W-MAE uses the standard VIT encoder to process the unmasked
patches, as shown in Figure~\ref{fig:architecture}. Subsequently,
all patches are fed into the W-MAE decoder in their original order.
The standard MAE learns representations by reconstructing an image
after masking out most of its pixels, and its decoder uses
self-attention for weight matrix computation over all input patches.
However, since meteorological images are approximately three times
larger than typical computer vision images, using the original MAE
decoder would result in a significant increase in computational
resources. Therefore, our W-MAE decoder splits the input
one-dimensional patch sequence into a two-dimensional patch matrix,
following the shape of the two-dimensional meteorological image. As
a result, our decoder only considers the relationships between patch
blocks along the second dimension when performing weight computation
via self-attention. Specifically, for an input image
$I\in\mathbb{R}^{H\times W\times C}$ with height, width, and channel
as $H$, $W$, and $C$, respectively, we divide it into patches with
each patch resolution set to $(P, P)$. As a result, we get $N =
HW/P^{2}$ one-dimensional patches. To ensure the original image
proportions are preserved, we then resize the individual $N$ patches
into a $(X, Y)$ two-dimensional patch matrix, where $X = H/P$ and $Y
= W/P $. Consequently, we can apply the standard self-attention
formula as shown below:
\begin{equation}
\begin{aligned}
\text{Attention}(\boldsymbol{Q},\boldsymbol{K},\boldsymbol{V})=\operatorname{softmax}(\frac{\boldsymbol{Q}\boldsymbol{K}^\top}{\sqrt{d}})\boldsymbol{V}.
\end{aligned}
\end{equation}
The query, key, and value vectors generated by the transformer block
are represented by $\textbf{Q, K}$, and $\textbf{V}$, respectively,
where the feature dimensionality of $\textbf{Q}$ and $\textbf{K}$ is
$d$. The original MAE decoder adopts a set of $N$ patch vectors as
values for $\textbf{Q, K}$, and $\textbf{V}$. We replace these with
an $(X, Y)$ two-dimensional patch vector matrix. To evaluate the
effectiveness of our improved structure, we employ a patch size of
$(32, 32)$ and measure the resulting performance. Notably, using our
W-MAE decoder achieves a 1.2 $\times$ speedup (for each training
epoch) compared with the vanilla MAE decoder, while reducing memory
overheads by up to 8\%. This significantly lowers the computational
resources demanded while still retaining high-quality pixel
reconstruction capabilities.

\section{Experiments}

In this section, we present pre-training and fine-tuning details for
W-MAE. As illustrated in Figure~\ref{fig:showcase}, our training
framework includes three stages, \textit{i.e.}, task-agnostic
pre-training, multi-variable forecast fine-tuning, and precipitation
forecast fine-tuning. We also provide visualization examples to
demonstrate the performance of W-MAE on the weather forecasting
tasks. Furthermore, we explore the inference efficiency and evaluate
the performance of ensemble forecast in multi-variable tasks.
Additionally, we conduct ablation studies to analyze the effects of
different pre-training settings.

\subsection{Implementation details of task-agnostic pre-training}
\label{sec:pretrain_implementation}

Given data samples from the ERA5 dataset, each sample contains
twenty atmospheric variables and is represented as a 20-channel
meteorological image. Our W-MAE first partitions meteorological
images into regular non-overlapping patches. Next, we randomly
sample patches to be masked, with a mask ratio of 0.75. The visible
patches are then fed into the W-MAE encoder, while the W-MAE decoder
reconstructs the missing pixels based on the visible ones. The mean
squared error between the reconstructed and original images is
computed in the pixel space and averaged over the masked patches.
There are some differences in model architecture for two dataset
division settings (see Section~\ref{sec:dataset} for details). For
clarity, we denote the models trained on two years of data and
thirty-seven years of data as W1 and W2, respectively. For W1, we
set the patch size as $16 \times 16$, while for W2, it is $8 \times
8$. The W1 model consists of encoders with depth=16, dim=768 and
decoders with depth=12, dim=512. The W2 model employs encoders with
depth=12, dim=768 and decoders with depth=6, dim=512. To reduce the
memory overhead and save on training time, we employ a smaller model
architecture for W2, given that the training data volume for W2 is
nearly 20 times that of W1. Furthermore, with a patch size of $8
\times 8$, the number of generated patches reaches the maximum limit
beyond which the self-attention network will be effective.
Therefore, in the W2 model, we replace the self-attention blocks
used in both the encoder and decoder with adaptive Fourier neural
operator (AFNO) blocks \cite{DBLP:journals/corr/abs-2111-13587}
obtained from FourCastNet. This process is equivalent to applying
the pre-training architecture of our W-MAE to FourCastNet. We employ
the AdamW optimizer \cite{DBLP:conf/iclr/LoshchilovH19} with two
momentum parameters $\beta_ 1$=0.9 and $\beta_2$=0.95, and set the
weight decay to 0.05. The pre-training process takes 600 epochs for
W1 and 200 epochs for W2. Note that W2 is currently still in
training and has not yet reached its optimal performance.

\subsection{Fine-tuning for multi-variable forecasting}

\begin{table}[t]
\centering \caption{Number of initial conditions used for computing
ACC and RMSE plots with the assumed temporal decorrelation time for
the variables $Z_{500}, T_{850}, T_{2m}, U_{10}, V_{10}, TP$.}
\resizebox{\textwidth}{!}{ }
\begin{tabular}{ccc}
\toprule
Variables & $N_f$ & $d$ (days)             \\
\midrule
$Z_{500}$ & 36 & 9 \\
$T_{850}$ & 36 & 9 \\
$T_{2m}$ & 40 & 9 \\
$U_{10}$ & 178 & 2 \\
$V_{10}$ & 178 & 2 \\
$TP$ & 180 & 2 \\
\bottomrule
\end{tabular}
\label{tab:initial_condition}
\end{table}

We perform multi-variable forecast fine-tuning on the ERA5 dataset,
with the task of predicting future states of meteorological
variables based on the meteorological data from the previous
time-step. We denote the modeled variables as a tensor $X(k\Delta
t)$, where $k$ represents the time index and $\Delta t$ is the
temporal spacing between consecutive time-steps in the dataset.
Throughout this work, we consider the ERA5 dataset as the
ground-truth and denote the true variables as $X_{true}(k\Delta t)$.
For simplicity, we omit $\Delta t$ in our notation, and $\Delta t$
is fixed at 6 hours. After pre-training, we fine-tune our
pre-trained W-MAE still using the ERA5 dataset to learn the mapping
from $X(k)$ to $X(k + 1)$. The fine-tuning process for
multi-variable forecasting takes a further 120 epochs and 100 epochs
for W1 and W2, respectively.

To evaluate the performance of our W-MAE model on multi-variable
forecasting, we perform autoregressive inference to predict the
future states of multiple meteorological variables. Specifically,
the fine-tuned W-MAE is initialized with $N_f$ different initial
conditions, taken from the testing split (\textit{i.e.}, data
samples from the year 2018). $N_f$ varies depending on the
forecasting days $d$, which differs for each forecast variable, as
shown in Table~\ref{tab:initial_condition}. Subsequently, the
fine-tuned W-MAE is allowed to run iteratively for $t$ time-steps to
generate future states at time-step $i + j\Delta t$. For each
forecast step, we evaluate the latitude-weighted Anomaly Correlation
Coefficient (ACC) and Root Mean Squared Error (RMSE)
\cite{DBLP:journals/ames/RaspDSWMT20} for all forecast variables.
The latitude-weighted ACC for a forecast variable $v$ at forecast
time-step $l$ is defined as follows:
\begin{equation}
\begin{aligned}
\tilde{\mathbf{X}}_{\mathrm{P/T}} =
\tilde{\mathbf{X}}_{\mathrm{pred/true}}(l)\left[v,m,n\right],
\end{aligned}
\end{equation}
\begin{equation}
\begin{aligned}
L(i)=\dfrac{\cos(\mathrm{lat}(m))}{\frac{1}{N_{\mathrm{lat}}}\sum_j^{N_{\mathrm{lat}}}\cos(\mathrm{lat}(m))},
\end{aligned}
\end{equation}
\begin{equation}
\begin{aligned}
\operatorname{ACC}(v,l)=
\dfrac{\sum_{m,n}L(m)\tilde{\mathbf{X}}_{\mathrm{P}}\tilde{\mathbf{X}}_{\mathrm{T}}}{\sqrt{\sum_{m,n}{L}(m)(\tilde{\mathbf{X}}_{\mathrm{P}})^2\sum_{m,n}{L}\left(m\right)(\tilde{\mathbf{X}}_{\mathrm{T}})^2}},
\end{aligned}
\end{equation}
where $\tilde{\mathbf{X}}_{\mathrm{pred/true}}(l)\left[v,m,n\right]$
represents the long-term-mean-subtracted value of predicted or true
variable $v$ at the location denoted by the grid coordinates $(m,n)$
at the forecast time-step $l$. The long-term mean of a variable
refers to the average value of that variable over a large number of
historical samples. The long-term mean-subtracted variables
$\tilde{\mathbf{X}}_{\mathrm{pred/true}}$ represent the anomalies of
those variables that are not captured by the long-term mean values.
$L(m)$ is the latitude weighting factor at the coordinate $m$ and
$\mathrm{lat}(m)$ denotes the latitude value. $N_{lat}$ provides the
total number of latitude locations in the dataset, which is used to
normalize the latitude weighting factor to maintain the overall
weight balance across all the locations on the grid. The
latitude-weighted RMSE for a forecast variable $v$ at forecast
time-step $l$ is defined as follows:
\begin{equation}
\begin{aligned}
{\mathbf{X}}_{\mathrm{P/T}} =
{\mathbf{X}}_{\mathrm{pred/true}}(l)\left[v,j,k\right],
\end{aligned}
\end{equation}
\begin{equation}
\begin{aligned}
\operatorname{RMSE}(v,l)=\sqrt{\frac{1}{N_{lat}N_{lon}}\sum_{j=1}^{N_{lat}}\sum_{k=1}^{N_{lon}}L(j)({\mathbf{X}}_{\mathrm{P}}-{\mathbf{X}}_{\mathrm{T}})^2},
\end{aligned}
\end{equation}
where ${\mathbf{X}}_{\mathrm{pred/true}}$ represents the value of
predicted or true variable $v$ at the location denoted by the grid
coordinates $(j, k)$ at the forecast time-step $l$. $N_{lat}$ and
$N_{lon}$ denote the total number of latitude locations and
longitude locations, respectively.

\begin{figure}[t]
\centering
\includegraphics[width=0.45\textwidth]{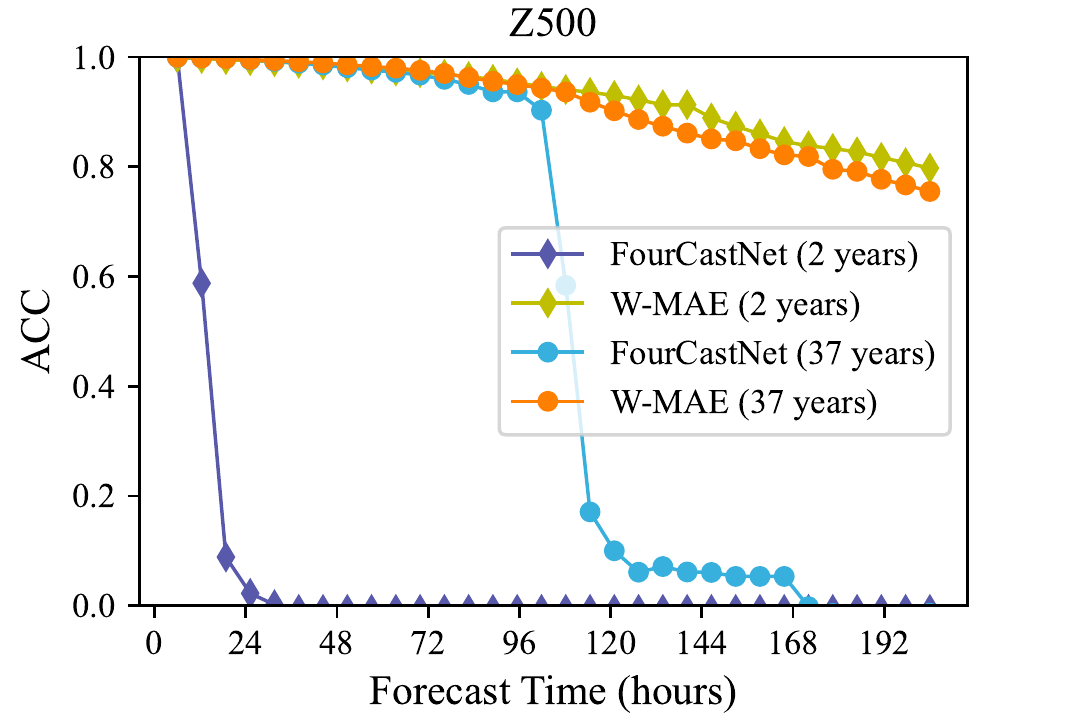}
\quad
\includegraphics[width=0.45\textwidth]{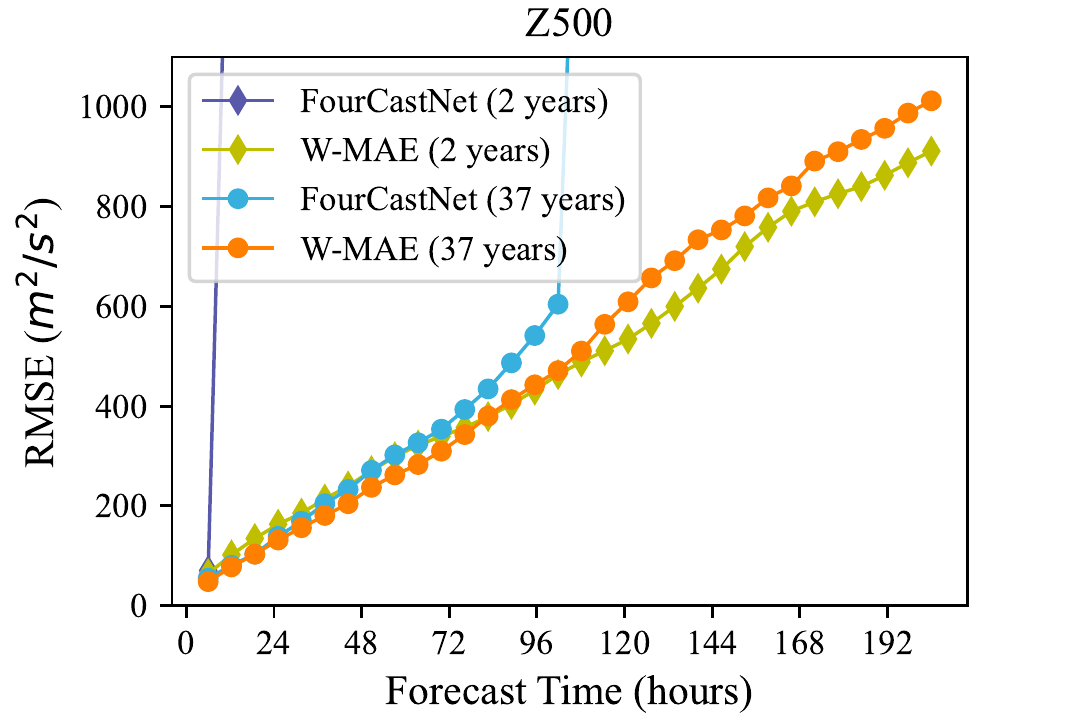}

\includegraphics[width=0.45\textwidth]{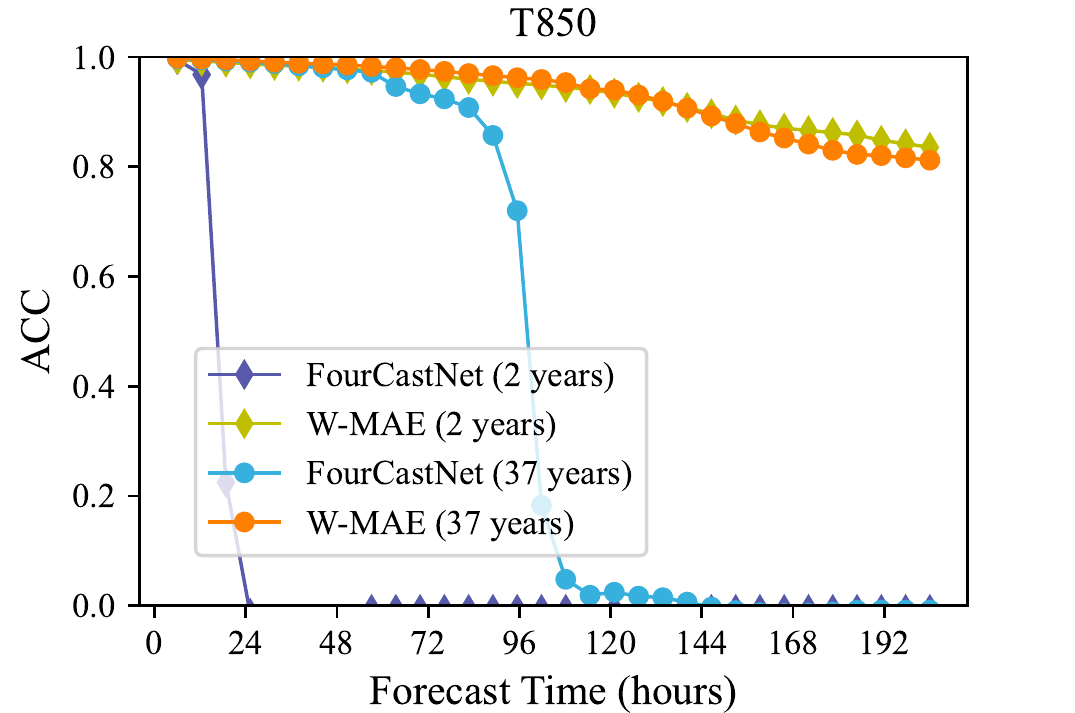}
\quad
\includegraphics[width=0.45\textwidth]{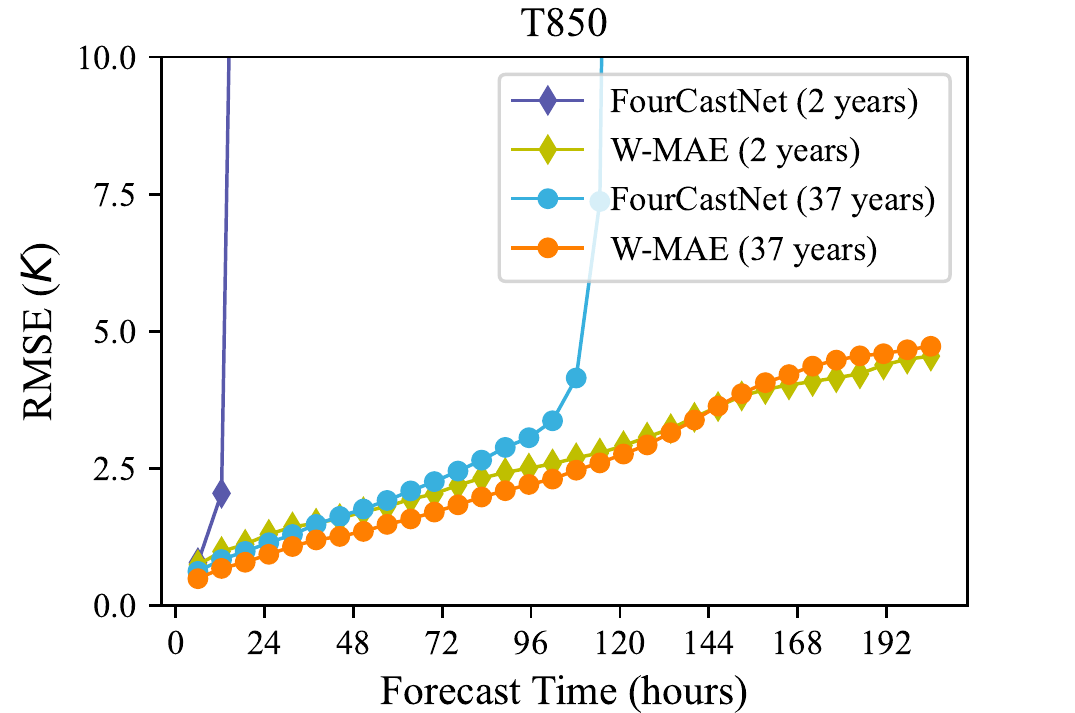}

\includegraphics[width=0.45\textwidth]{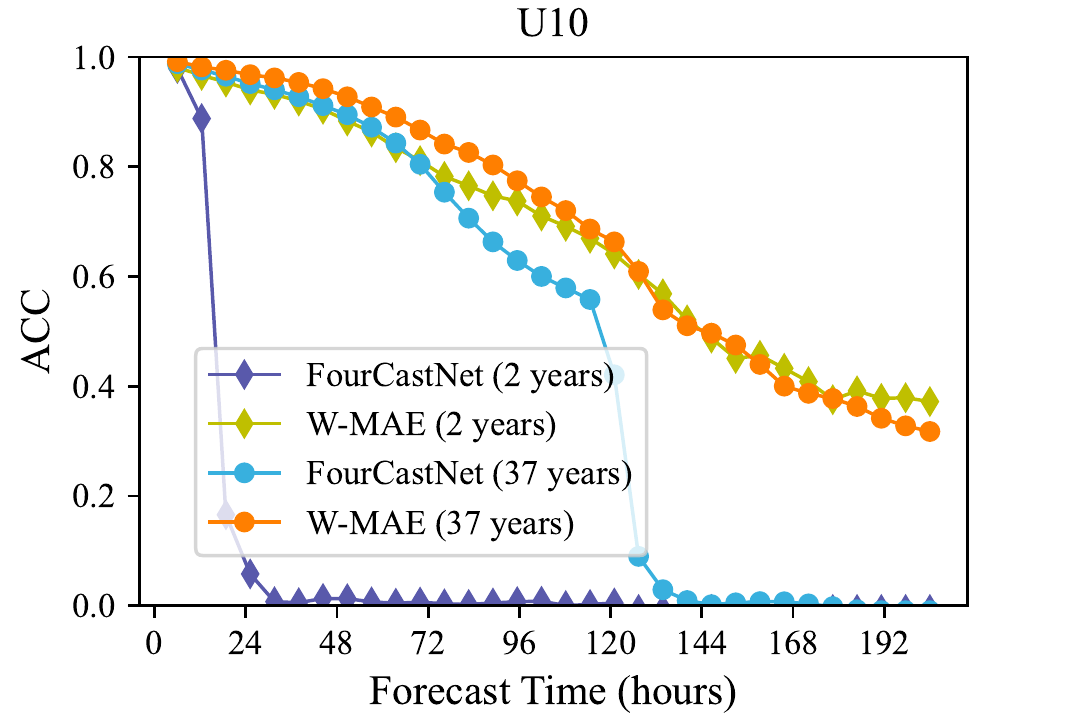}
\quad
\includegraphics[width=0.45\textwidth]{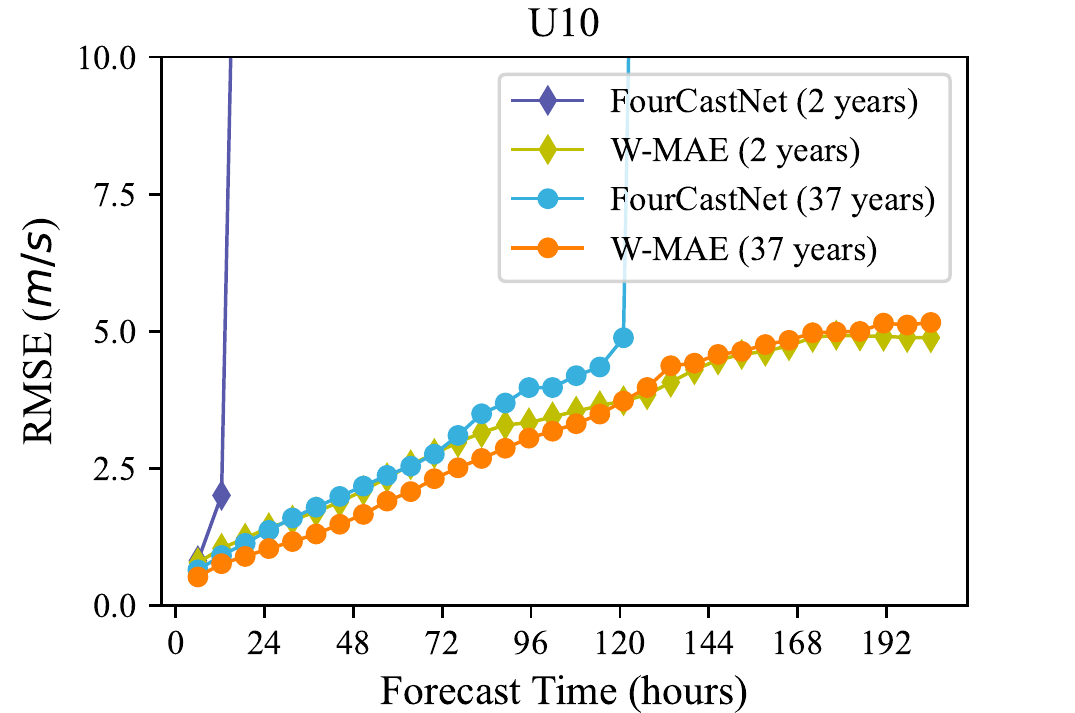}

\includegraphics[width=0.45\textwidth]{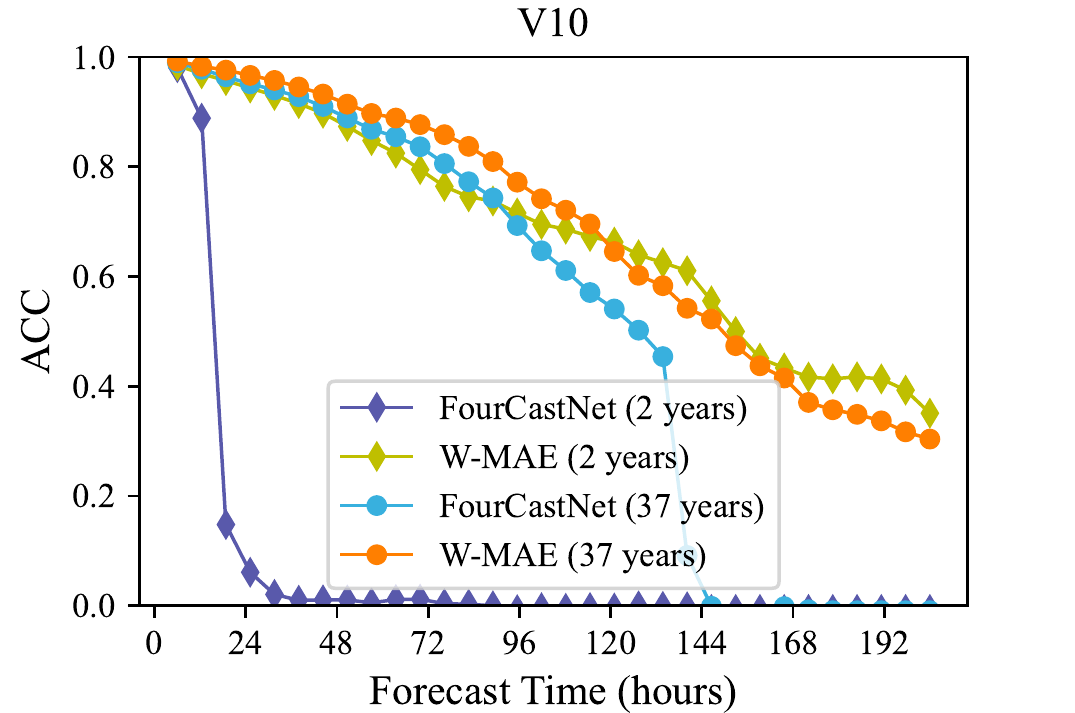}
\quad
\includegraphics[width=0.45\textwidth]{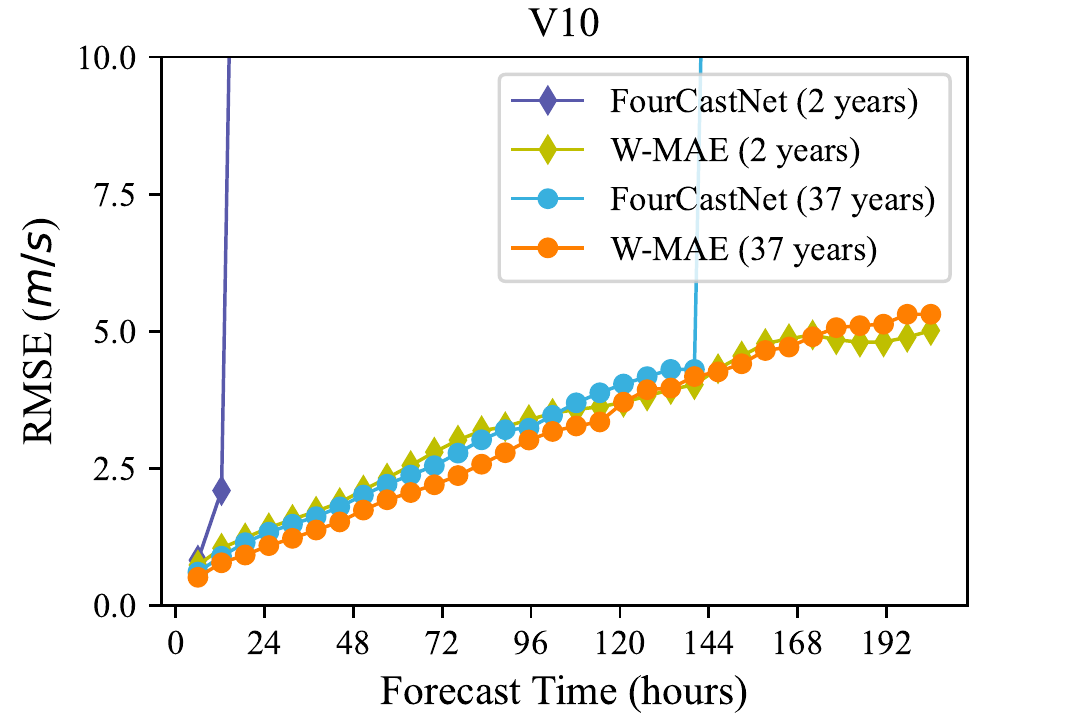}
\caption{The latitude-weighted ACC and RMSE curves for our W-MAE
forecasts and the corresponding matched FourCastNet forecasts at a
fixed initial condition in the testing dataset corresponding to the
calendar year 2018 for the variables, including $Z_{500}$,
$T_{850}$, $V_{10}$, and $U_{10}$.} \label{fig:comparison_meteorological_variables}
\end{figure}

We report the mean ACC and RMSE for each of the variables at each
forecast time-step and compare them with the corresponding
FourCastNet forecasts that use time-matched initial conditions. We
present the performance comparisons of W-MAE and FourCastNet under
two different training data durations, namely, two years and
thirty-seven years.
Figure~\ref{fig:comparison_meteorological_variables} displays the
latitude-weighted ACC and RMSE values for our W-MAE model forecasts
(indicated by the red line with markers) and the corresponding
FourCastNet forecasts (indicated by the blue line with markers).

As depicted in Figure~\ref{fig:comparison_meteorological_variables},
with two years of training data, the W-MAE pre-trained delivers
stable and satisfactory performance in forecasting the future states
of multiple meteorological variables. In contrast, FourCastNet
exhibits a rapid drop in performance at very short lead times and
becomes unable to predict beyond 24 hours. The reason for this is
that our W-MAE model, through pre-training, can learn the basic
structure and patterns within the data from the limited training
data. Subsequently, the pre-trained W-MAE model utilizes the learned
fundamental data features which serve as prior knowledge for the
model, making it more robust in downstream task fine-tuning.
However, traditional models in the case of inadequate training data
encounter two limitations: (1) limited capacity to capture latent
patterns and features, leading to overfitting or poor performance,
and (2) more prone to be affected by noise and outliers, reducing
predictive robustness. These limitations may account for the
underperformance of FourCastNet.

Under the setting of thirty-seven years of training data, our W-MAE
continues to outperform FourCastNet, maintaining stable prediction
performance. It is important to note that FourCastNet incorporates a
distinct approach to mitigate accumulation errors (see
\cite{DBLP:journals/corr/abs-2202-11214} for more details),
necessitating a second round of training for multi-variable
forecasting. To ensure fairness and enable a straightforward
comparison, we adopt a first-round-training model of FourCastNet
with a training epoch of 150 (matching the count mentioned in
FourCastNet). It is observed that the performance of FourCastNet,
which lacks the integration of a tailored training process for error
accumulation suppression, experiences a significant decline beyond a
certain forecast lead time. These findings underscore the
superiority of our W-MAE framework in maintaining performance
stability and robustness of results. Additionally, in
Figure~\ref{fig:comparison_meteorological_variables}, we observe
that beyond a certain lead time (approximately 130 hours), the
performance of our W-MAE model with thirty-seven years of training
data is not as strong as that achieved with two years of training
data. This may be attributable to the fact that our W-MAE model with
thirty-seven years of training data has not yet converged to its
optimal performance. As mentioned in
Section~\ref{sec:pretrain_implementation}, our W-MAE model with two
years of training data undergoes 600 epochs of pre-training, while
the model with thirty-seven years of training data is only
pre-trained for 200 epochs.

Moreover, we conduct an analysis of the time overhead for
fine-tuning our pre-trained W-MAE model in multi-variable
forecasting tasks. The fine-tuning process takes 100 epochs, with
each epoch lasting an average of 63 minutes on 8 NVIDIA Tesla A800
GPU, resulting in a total fine-tuning time of nearly 105 hours. In
comparison, FourCastNet requires 150 epochs of training, averaging
55 minutes per epoch on the same platform resulting in a total
training time of around 137 hours. The results indicate that our
W-MAE achieves a nearly 23\% speedup compared with FourCastNet. This
demonstrates that by leveraging the powerful initial parameters
provided by pre-training, our W-MAE can achieve fast convergence
during the fine-tuning process, which saves training time and
computational resources.

\begin{figure}[t]
\centering
\includegraphics[width=\textwidth]{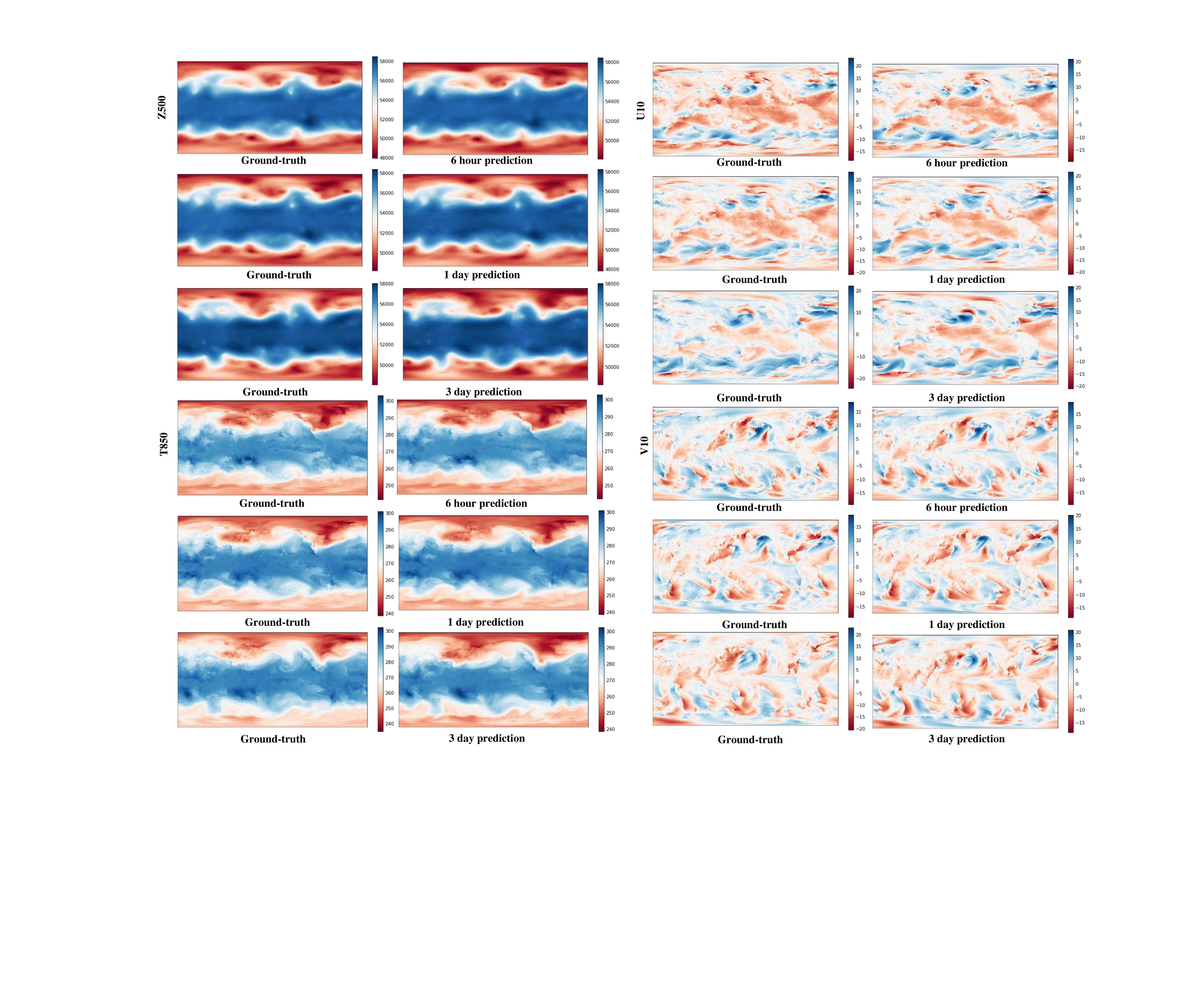}
\caption{Visualization examples of future state prediction for
meteorological variables, including $Z_{500}$, $T_{850}$, $V_{10}$,
and $U_{10}$.} \label{fig:longStepPredict}
\end{figure}

\subsection{Fine-tuning for precipitation forecasting}

\begin{figure}[t]
\centering
\includegraphics[width=0.45\textwidth]{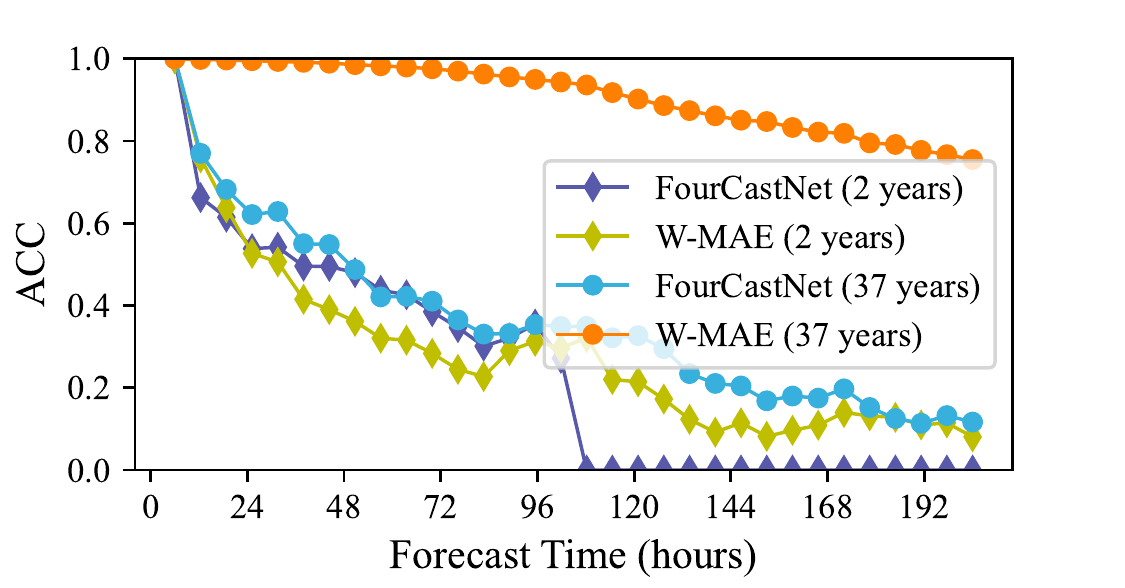}
\quad
\includegraphics[width=0.43\textwidth]{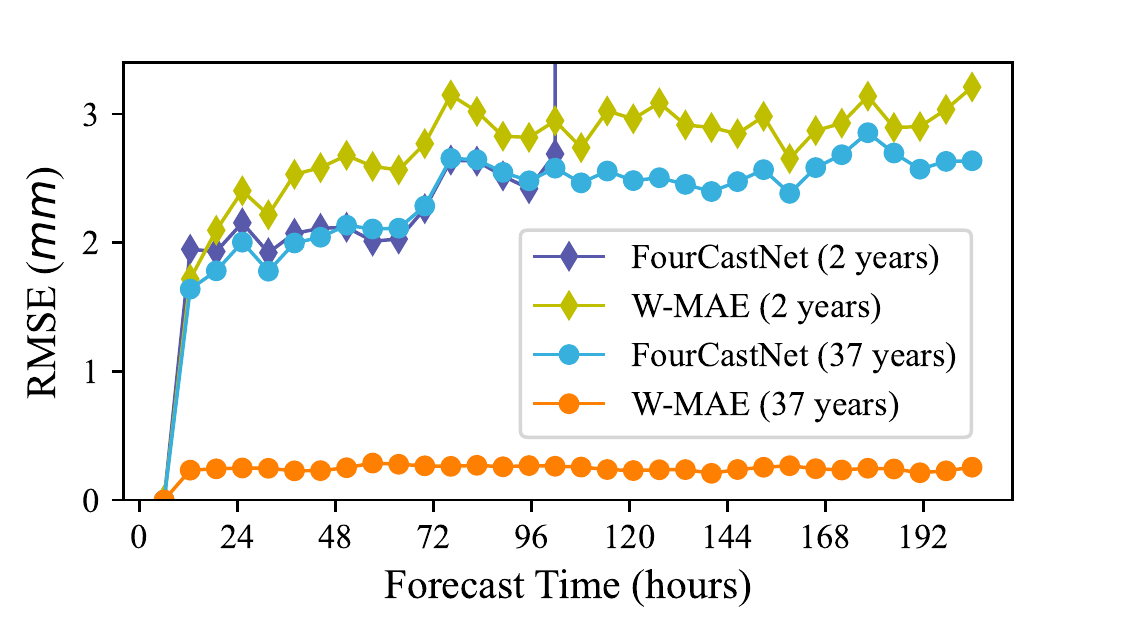}
\caption{Latitude weighted ACC and RMSE curves for our W-MAE
forecasts and the corresponding matched FourCastNet forecasts at a
fixed initial condition in the testing dataset corresponding to the
calendar year 2018 for TP. Please note that the results of
``FourCastNet (37 years)'' are obtained using the checkpoint file
released by Pathak \etal \cite{DBLP:journals/corr/abs-2202-11214},
representing the optimal results after two rounds of training on
FourCastNet.} \label{fig: ComparisonTP16}
\end{figure}

Total Precipitation (TP) in the ERA5 dataset is a variable that
represents the cumulative liquid and frozen water that falls to the
Earth's surface through rainfall and snowfall. Compared with other
meteorological variables, TP presents sparser spatial
characteristics. It is calculated from other variables, requiring an
intermediate calculation in the derivation process, making it more
challenging. For the precipitation forecasting task, we aim to
predict the cumulative total precipitation in the next 6 hours using
multiple meteorological variables from the previous time-step. We
compare the performance of our W-MAE model and FourCastNet using
both ACC and RMSE as the evaluation metrics.

Experimental results for different training data settings,
\textit{i.e.} using two years or thirty-seven years of training
data, are shown in Figure~\ref{fig: ComparisonTP16}\footnote{Note
that, only a portion of FourCastNet's prediction results are
displayed in Figure~\ref{fig: ComparisonTP16}. This is because the
inability of FourCastNet to make predictions at longer lead times of
more than 108 hours for two years of training data. Specifically,
our experimental results show that in precipitation forecasting,
FourCastNet produces considerable prediction errors beyond a certain
lead time, resulting in NAN (not a number) predictions that render
the results infeasible.}. When using two years of training data, our
W-MAE model outperforms FourCastNet in predicting the total
precipitation at the 6-hour result. When the training data spans
thirty-seven years, at the first forecast time-step (\textit{i.e.},
with a forecast lead time of 6 hours), our model achieves a 20\%
improvement in terms of the ACC metric compared with FourCastNet
(0.996 vs. 0.804). At the 20$th$ forecast time-step (\textit{i.e.},
with a forecast lead time of 5 days), our model significantly
outperforms FourCastNet (0.886 vs. 0.234). It is worth noting that
despite incorporating the error accumulation suppression
optimization techniques and undergoing a second round of training
(see \cite{DBLP:journals/corr/abs-2202-11214} for detailed training
process), the yielding FourCastNet model still exhibits inferior
performance in precipitation forecasting compared to our W-MAE
model. Moreover, concerning the time costs associated with
precipitation forecasting tasks, our W-MAE framework outperforms
FourCastNet, with a total training time of 20.6 hours and 23 hours,
respectively. These observations suggest that our W-MAE not only
achieves superior performance but also reduces the training time,
resulting in efficient model training.

\begin{figure}[t]
\centering
\includegraphics[width=0.45\textwidth]{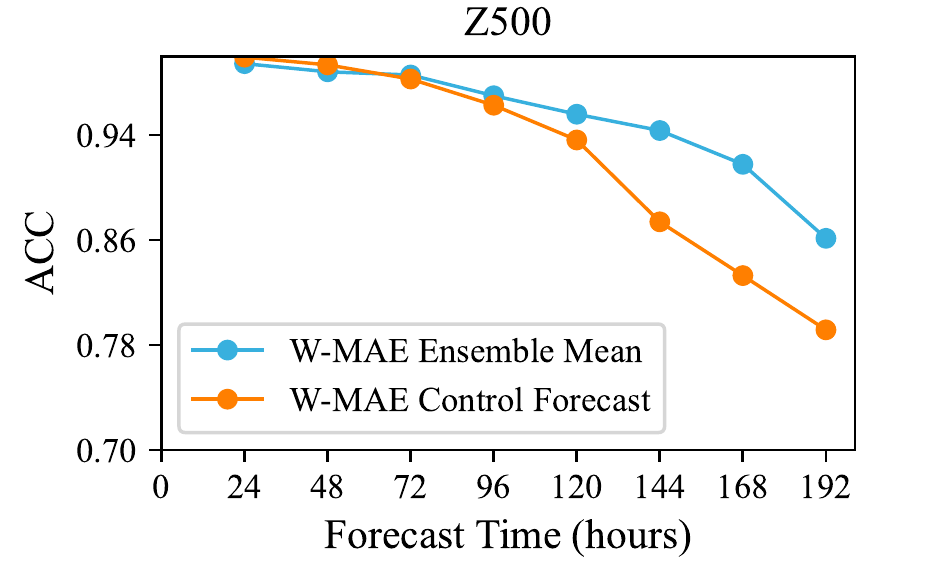}
\quad
\includegraphics[width=0.45\textwidth]{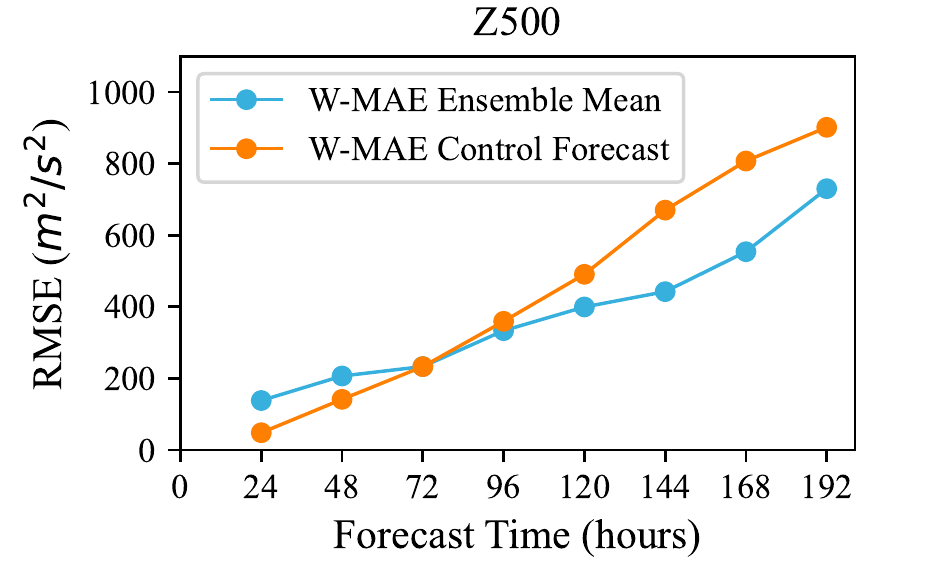}
\caption{Comparison of control forecast and ensemble forecast
results of W-MAE. We display the latitude-weighted RMSE and ACC for
$Z_{500}$.} \label{fig:ensemble}
\end{figure}

\subsection{Efficiency of ensemble forecast}

The methodology of ensemble forecast mainly involves adding noise to
perturb initial weather states and observing the change in forecast
results. In this paper, we follow FourCastNet
\cite{DBLP:journals/corr/abs-2202-11214} to use Gaussian noise to
perturb the initial conditions and generate an ensemble of 100
perturbed initial condition sets (represented as $E$=100). The
perturbations are scaled by a factor $\sigma$=0.3. We apply
large-member ensemble on multi-variable forecasting tasks for
$Z_{500}$. As shown in Figure~\ref{fig:ensemble}, we compare control
forecast (\textit{i.e.}, the unperturbed forecasts) with the mean of
a 100-member ensemble forecast using W-MAE. For the short-term
forecast time, such as for 1 day, 100-member ensemble forecasts
exhibit a slightly lower accuracy than control forecasts by a single
model. However, when predicting beyond 5 days, the accuracy of
100-member ensemble forecasts displays a marked improvement compared
with control forecasts. This observation affirms that the
large-member ensemble forecast is valuable when the precision of
individual models is compromised. Moreover, our W-MAE takes an
average of 320 ms to infer a 24-hour forecast on one NVIDIA Tesla
A800 GPU, which is orders of magnitude faster than traditional NWP
methods in ensemble forecast. The results prove the efficiency of
our W-MAE on ensemble forecast.

\subsection{Ablation study}

\begin{figure}[t]
\centering
\includegraphics[width=0.45\textwidth]{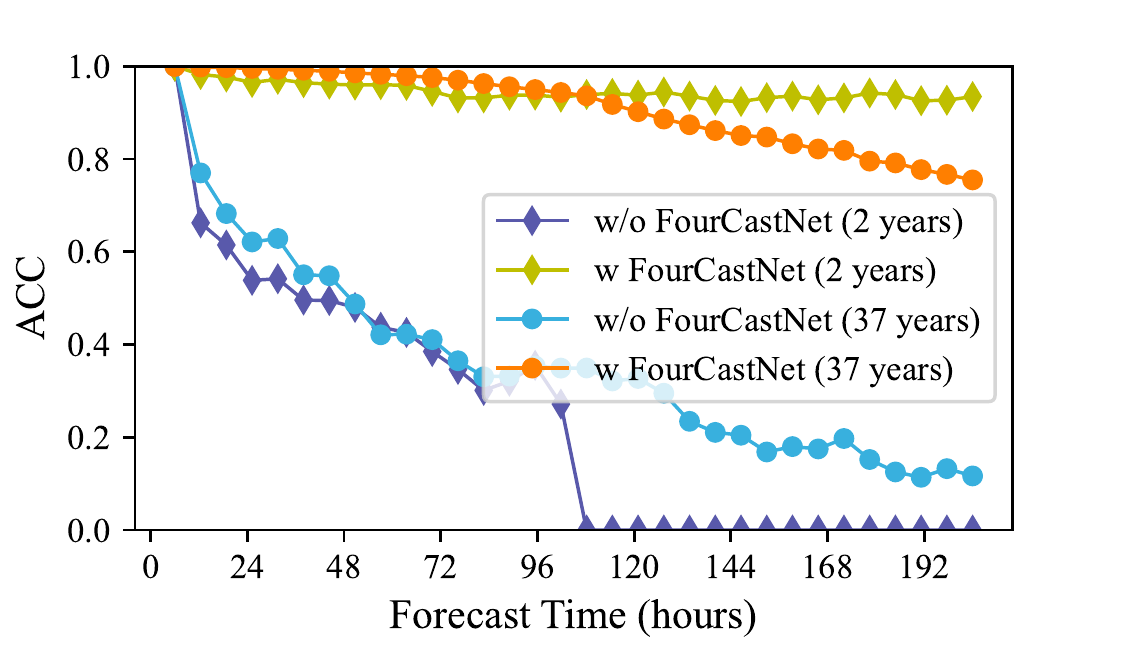}
\quad
\includegraphics[width=0.45\textwidth]{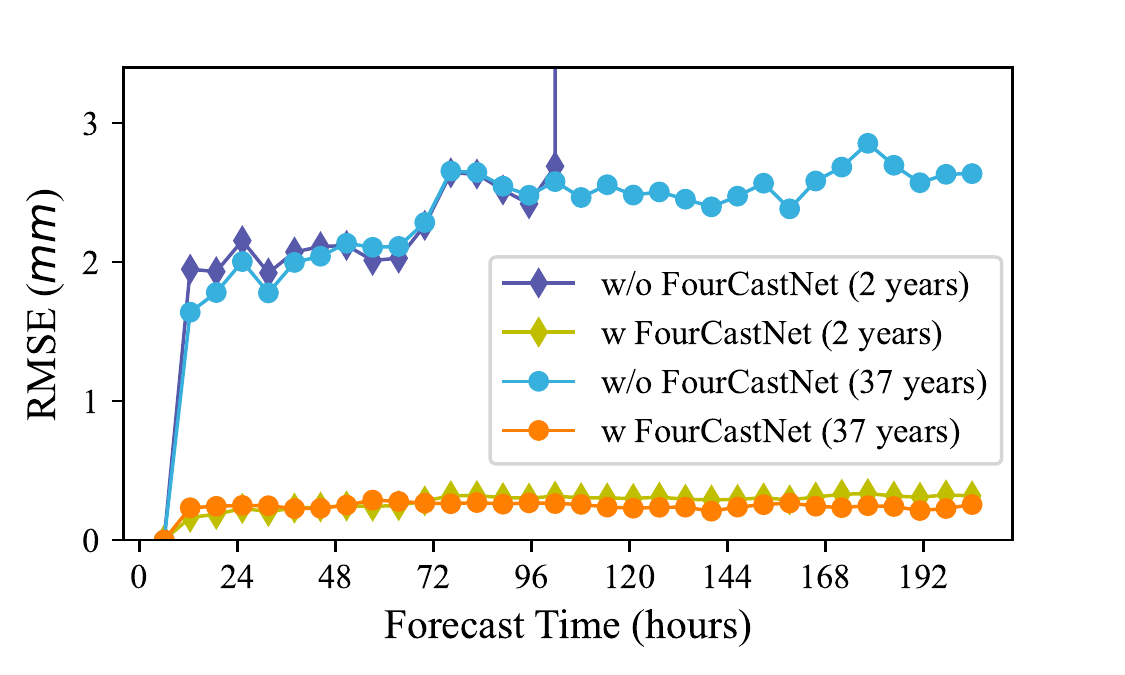}
\caption{Latitude weighted ACC and RMSE curves for our W-MAE
forecasts and the corresponding matched FourCastNet forecasts at a
fixed initial condition in the testing dataset corresponding to the
calendar year 2018 for TP.} \label{fig:ComparisonTP8}
\end{figure}

\textbf{Effect of pre-training.} To fully validate the effectiveness
of our method, we apply our pre-training technique to FourCastNet.
During the pre-training stage, we set the patch size to $8 \times 8$
(same as that used in FourCastNet), the mask ratio to 0.75, and
trained the model for 250 epochs. Then, we fine-tune the pre-trained
FourCastNet for the precipitation forecasting task. We compared the
performance of the pre-trained FourCastNet with that of the vanilla
FourCastNet without pre-training. Our ablation results, as shown in
Figure~\ref{fig:ComparisonTP8}, indicate that the pre-trained
FourCastNet outperformed FourCastNet without pre-training by nearly
30\% in precipitation forecasting when both were trained on two
years of data. Moreover, the pre-trained FourCastnet outperformed
FourCastNet without pre-training, which was trained on thirty-seven
years of data, by nearly 20\%. Furthermore, in both training data
settings, \textit{i.e.}, using two years and thirty-seven years of
training data, FourCastnet with pre-training shows a significant
performance improvement compared with that without pre-training.
This suggests that applying pre-training to weather forecasting is
promising.

\begin{figure}[t]
\centering
\includegraphics[width=0.45\textwidth]{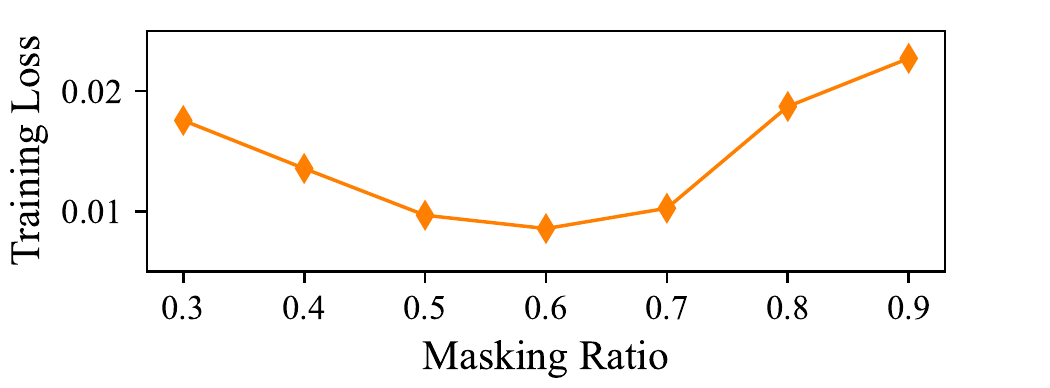}
\quad
\includegraphics[width=0.45\textwidth]{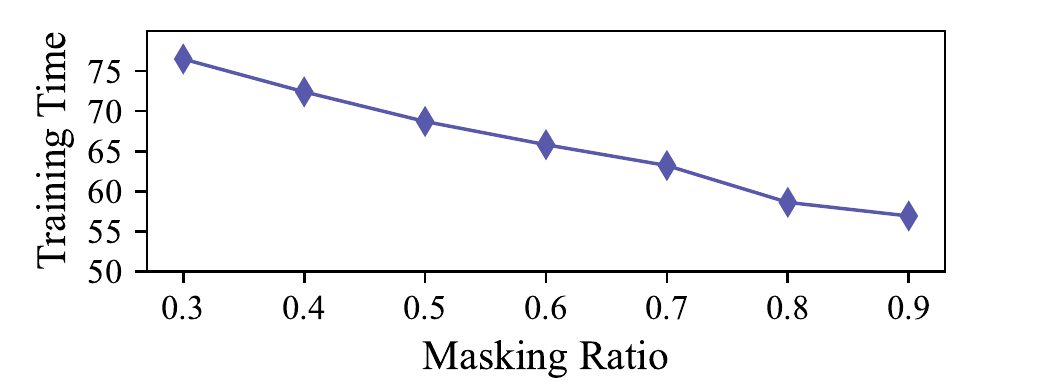}
\caption{Effect of masking ratio on training loss and training time.
The optimal masking ratio was found to be approximately 60\%.}
\label{fig:maskRatio_loss}
\end{figure}

\begin{figure}[t]
\centering
\includegraphics[width=0.82\textwidth]{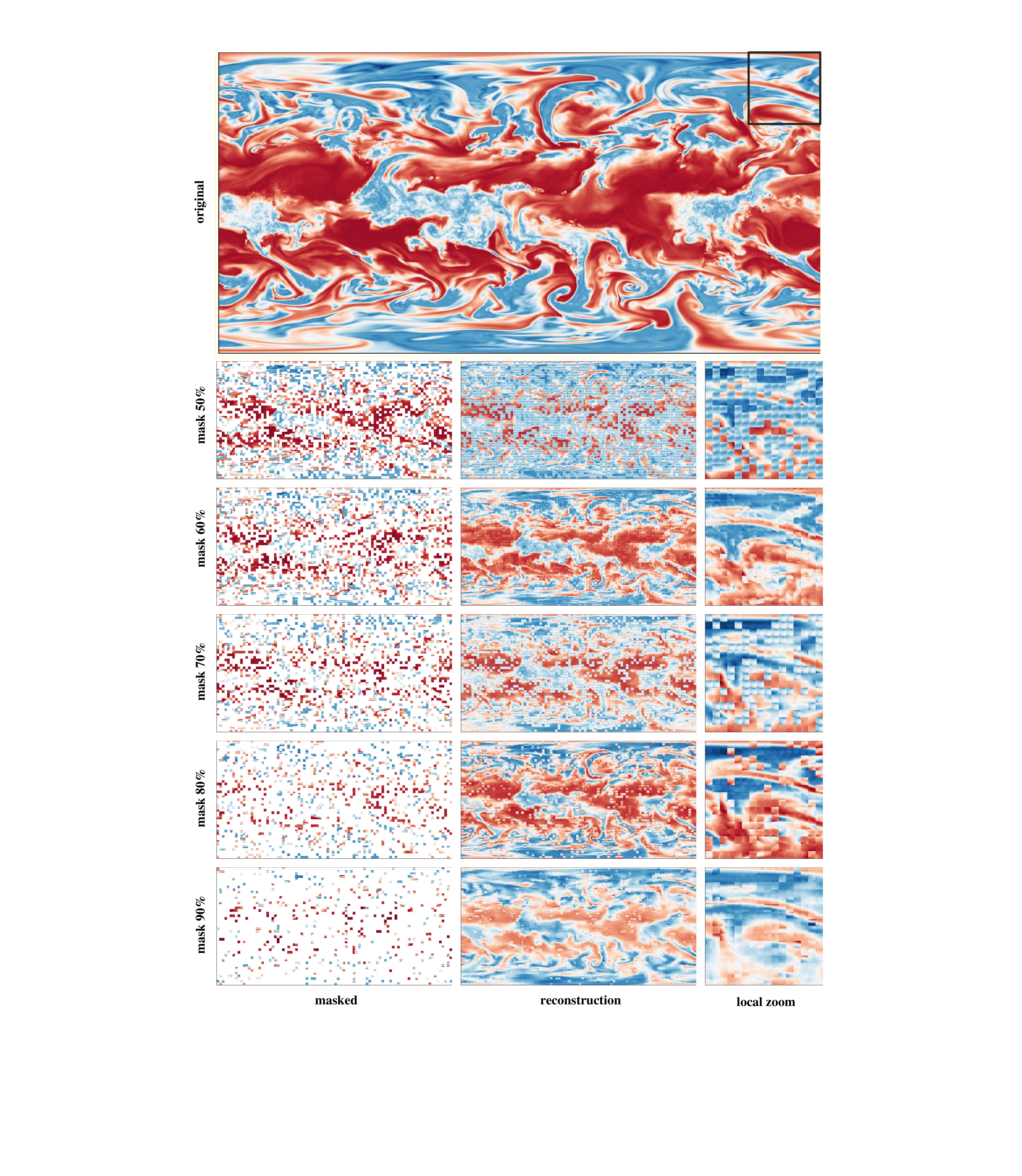}
\caption{Visualization of the pixel reconstruction results under
varying mask ratio condition. The `local zoom' column displays the
magnified results of the local area wrapped by the black box in the
original images.} \label{fig:maskRatio}
\end{figure}

\textbf{Effect of masking ratio.} Given the disparity in information
density between meteorological images and images in computer vision,
we conducted an investigation on the effect of mask ratios on
meteorological image reconstruction. Specifically, the training loss
under different mask ratios are compared and we set the maximum
pre-training epoch to 1000. Figure~\ref{fig:maskRatio_loss} displays
the effect of masking ratios on pixel reconstruction, with the
optimal ratio found at approximately 0.6. However, when considering
the time overhead of pre-training, we discover that setting the mask
ratio to 0.75, compared with 0.6, saves 8\% of the computational
memory usage and reduces the training time by nearly 4 minutes per
epoch. Moreover, the performance of the model at a 0.75 mask ratio
can be comparable to that of 0.6. Therefore, to strike a balance
between computational efficiency and model performance, we select a
mask ratio of 0.75 for model pre-training. Furthermore,
Figure~\ref{fig:maskRatio} visualizes the pixel reconstruction
results under varying mask ratio conditions.

\section{Conclusion}

In this paper, we introduce self-supervised pre-training techniques
to the weather forecasting domain, and propose a pre-trained Weather
model with Masked AutoEncoder named W-MAE. Our pre-trained W-MAE
exhibits stable forecasting results and outperforms the baseline
model in longer forecast time-steps. This demonstrates the
feasibility and potential of implementing pre-training techniques in
weather forecasting. Our study also provides insights for modeling
longer-term dependencies (ranging from a month to a year) in the
tasks of climate forecasting \cite{DBLP:journals/jc/MartinMSBIRK10,
DBLP:journals/rms/Hoskins13}. We hope that our work will inspire
future research on the application of pre-training techniques to a
wider range of weather and climate forecasting tasks.

\backmatter

\bmhead{Acknowledgments} The authors gratefully acknowledge the
support of MindSpore, CANN (Compute Architecture for Neural
Networks) and Ascend AI Processor used for this research. The
computing infrastructure used in our work is powered by Chengdu
Intelligent Computing Center.

\section*{Declarations}


\begin{itemize}
\item \textbf{Funding} This work was supported by the National Natural Science Foundation
of China (No. 62276047 and No. 42275009) and CAAI-Huawei MindSpore
Open Fund.
\item \textbf{Conflict of interest} The authors have no financial or non-financial interests to disclose.
\item \textbf{Ethics approval and Consent to participate} The authors declare that this research did not require Ethics
approval or Consent to participate since no experiments involving
humans or animals have been conducted.
\item \textbf{Consent for publication} The authors of this manuscript all consent to its publication.
\item \textbf{Availability of data and materials and Code availability} The code and data are available at
\url{https://github.com/Gufrannn/W-MAE}.
\item \textbf{Authors' contributions}: Conceptualization: Xin Man,
Chenghong Zhang; Methodology: Xin Man, Jin Feng, Changyu Li, Jie
Shao; Formal analysis and investigation: Xin Man, Chenghong Zhang,
Jin Feng, Changyu Li; Writing - original draft preparation: Xin Man;
Writing - review and editing: Jin Feng, Jie Shao; Funding
acquisition: Jin Feng, Jie Shao.
\end{itemize}


\bigskip


\bibliography{sn-bibliography}


\end{document}